\pdfoutput=1
\documentclass[10pt,twocolumn]{article}

\usepackage[margin=0.72in,columnsep=0.26in]{geometry}
\usepackage[T1]{fontenc}
\usepackage[utf8]{inputenc}
\usepackage{lmodern}
\usepackage{microtype}
\usepackage{amsmath,amssymb}
\usepackage{graphicx}
\usepackage{booktabs}
\usepackage{multirow}
\usepackage{xcolor}
\usepackage{hyperref}
\usepackage[numbers,sort&compress]{natbib}
\usepackage{caption}
\usepackage{enumitem}

\hypersetup{
  colorlinks=true,
  linkcolor=blue!55!black,
  citecolor=blue!55!black,
  urlcolor=teal!60!black
}

\setlist[itemize]{leftmargin=1.2em,itemsep=2pt,topsep=3pt}
\setlist[enumerate]{leftmargin=1.4em,itemsep=2pt,topsep=3pt}

\title{\textbf{The Resolution Blind Spot: How Proxy Metrics and LLM-as-a-Judge Mask Forgetting and Reject Pareto-Superior Models in Low-Resource 7B Fine-Tuning}}

\author{
  Yuyang Zhang \qquad Baosong Fan\\
  Chengdu University of Technology\\
  \texttt{Code \& Audit Suite: sft-audit}
}

\date{}

\begin{document}
\maketitle

\begin{abstract}
Practitioners adapting 7B-scale open-weight language models on consumer hardware (8\,GB VRAM) routinely iterate through synthetic data generation, 4-bit QLoRA fine-tuning, and pairwise LLM-as-a-Judge evaluation.
Through a forensic audit of an end-to-end 10-checkpoint domain-adaptation pipeline for Chinese computer science education (\texttt{Qwen2.5-7B-Instruct}) and an accompanying retrieval upgrade (\texttt{bge-small} to \texttt{bge-large}), we uncover a compound \textbf{Resolution Blind Spot} that causes both false optimism and false rejection:
(1)~Standard open-ended proxies (validation loss, character F1, and single-pass judge scores) reward textbook style mimicry while masking statistically significant catastrophic forgetting on 1,359 uncontaminated, zero-noise objective CS/AI questions from C-Eval and CMMLU: all six chained incremental $r{=}8$, effective-batch-1 adapters lose $-2.43$ to $-3.75$ percentage points (pp) relative to the base model ($70.86\% \to 67.11\%$--$68.43\%$, exact McNemar $p \le 0.039$, with up to $-12.3$\,pp loss on Computer Networks) and suffer up to $-14.9$\,pp format-compliance drift.
(2)~When structural fixes---fresh-start retraining, doubled LoRA rank ($r{=}16$), larger effective batch size ($8$), and a grounded LLM-verified data gate (2,794 verified pairs)---completely eliminate this forgetting ($70.79\%$ vs.\ base $70.86\%$, $p{=}1.000$; $+2.87$\,pp over the deployed model, $p{=}0.0084$) and cut representational prediction flips by $57\%$ ($230 \to 99$), a 144-item position-swapped LLM judge returns an exact $61{:}61$ ($p{=}1.000$) tie, leading the engineer to \emph{discard the Pareto-superior model}.
(3)~Exact binomial power analysis explains why: as candidates converge, judge position-flip rates climb from $11.8\%$ to $21.5\%$, pushing the 80\%-power Minimum Detectable Effect (MDE) of a 144-item benchmark to $13.1$--$13.7$\,pp (leaving $<18\%$ power for realistic $+5$\,pp gains, which instead require $N \approx 983$ prompts).
(4)~A parallel retrieval audit reveals a similar trap: naive self-prefix testing falsely ranks \texttt{bge-large} $-9.7$\,pp below \texttt{bge-small}, while a 934-query natural benchmark shows a statistically null gap ($88.76\%$ vs.\ $87.90\%$, McNemar $p{=}0.5195$) at $14.7\times$--$15.9\times$ compute overhead.
We release \texttt{sft-audit}, a lightweight CPU-only four-gate auditing toolkit to prevent these failure modes in resource-constrained workflows.
\end{abstract}

\section{Introduction}
Low-resource domain adaptation of 7B--8B instruction-tuned models via 4-bit QLoRA~\cite{dettmers2023qlora,hu2022lora} has become the default recipe for academic labs, startups, and vertical-domain engineering teams operating on single consumer GPUs.
Because domain experts rarely have budget to annotate thousands of gold open-ended responses, a widespread iterative loop has emerged: synthesize domain QA pairs with a frontier LLM, fine-tune a local 7B adapter, evaluate candidates with surface overlap (ROUGE/char-F1) or pairwise LLM-as-a-Judge~\cite{zheng2023judging}, diagnose failure cases, synthesize targeted patches, and repeat.

In practice, this loop frequently stalls after one or two iterations.
When subsequent rounds yield non-significant pairwise Arena results, practitioners face a fundamental ambiguity: \emph{did training genuinely hit a model-capacity or synthetic-data ceiling, or did the evaluation instrument lose resolution?}

We answer this question through a comprehensive empirical autopsy of a real, multi-round Chinese computer science adaptation project built on \texttt{Qwen2.5-7B-Instruct}~\cite{qwen25} on an 8\,GB RTX~5060 Laptop GPU.
Across 10 preserved checkpoints spanning six chained incremental runs and two fresh-start structural redesigns, plus 7 pairwise position-swapped Arena evaluations and a 934-query RAG embedding migration, we pair the original open-ended logs with a newly executed, strictly uncontaminated 1,359-question objective ground-truth anchor drawn from C-Eval~\cite{huang2023ceval} and CMMLU~\cite{li2024cmmlu}.

Our audit exposes four actionable findings:
\begin{enumerate}
  \item \textbf{Proxy and Retrieval Inversions.} Validation loss diverges from generation quality after epoch~1; focused weak-point SFT achieves the highest character F1 ($0.5435$) while inflicting the worst objective knowledge loss ($-3.75$\,pp, $p{=}0.0012$); and self-prefix embedding tests invert the ranking of \texttt{bge-large} vs.\ \texttt{bge-small} by $-9.7$\,pp.
  \item \textbf{Asymmetric Dilution and Leakage.} Targeted improvements ($22{:}9$, $p{=}0.029$) vanish under full-benchmark averaging ($69{:}71$, $p{=}0.933$), while generating error-correction data from evaluation misses injects $22\%$ item overlap ($12$ pairs at Jaccard $\ge 0.50$, $3$ verbatim matches), inflating apparent win rates by $+6$\,pp relative to a strictly decontaminated out-of-domain test set.
  \item \textbf{Hidden Forgetting and Rejection of a Pareto-Superior Checkpoint.} Chained $r{=}8$, effective-batch-1 adapters silently erode objective CS accuracy ($70.86\% \to 67.92\%$ for the deployed checkpoint, $p{=}0.010$) and MCQ format probability mass ($100.0\% \to 85.1\%$--$90.8\%$). Retraining from scratch with $r{=}16$, effective batch~$8$, and a grounded verification gate (\texttt{verified}) completely eliminates forgetting ($70.79\%$, $p{=}1.000$ vs.\ base; $+2.87$\,pp over deployed \texttt{corr}, $p{=}0.0084$) and restores format compliance ($99.6\%$)---yet was \emph{rejected} as a $61{:}61$ tie by the 144-item pairwise LLM judge.
  \item \textbf{Formalizing the Judge Resolution Ceiling.} We derive exact binomial Minimum Detectable Effects (MDE) under position-swapped judging with indecisive rate $\eta = \rho + \tau$ (position flips plus ties). Because $\rho$ rises from $11.8\%$ to $21.5\%$ near model parity, $N{=}144$ prompts only resolve $\ge 13.1$\,pp win-rate gaps at $80\%$ power, turning fine-grained LLM-judge hill-climbing into a random walk.
\end{enumerate}

\section{Experimental Setup and Iterative Trajectory}
\label{sec:setup}

\paragraph{Hardware and Base Model.}
All fine-tuning and inference experiments run on a single workstation with an Intel Core i7-14650HX CPU, 14\,GB RAM, and an NVIDIA RTX~5060 Laptop GPU (8\,GB VRAM).
The base model is \texttt{Qwen2.5-7B-Instruct} loaded in 4-bit NF4 double quantization with \texttt{bfloat16} compute.
Fitting 7B QLoRA training inside 8\,GB VRAM required disabling \texttt{transformers}~5.x caching allocator warmup (which pre-reserves $3.16$\,GiB), keeping only LayerNorm weights in \texttt{fp32} rather than upcasting the 152k-token embedding and LM head ($+2.03$\,GiB), switching from \texttt{paged\_adamw\_8bit} to Adafactor, and segmented checkpoint restarts to bound slow allocator growth.

\paragraph{Ten-Checkpoint Adaptation Lineage.}
Table~\ref{tab:checkpoints} summarizes the 10 evaluated models spanning the full iterative lifecycle:
(i)~the unadapted \texttt{base} model;
(ii)~seven \textbf{chained incremental} $r{=}8, \alpha{=}16$ runs initialized sequentially from an initial 1-epoch adapter (\texttt{init}) with effective batch size~$1$: 3-epoch textbook/curriculum continuation (\texttt{continue}, 3,436 pairs), focused pure-weakness SFT (\texttt{weak\_focus}, $71\times 3$ steps), 1:1 mixed weakness SFT (\texttt{weak\_1to1}), error-driven correction SFT (\texttt{corr}, deployed as production), contrastive concept-pair SFT (\texttt{concept\_v2}), and decontaminated out-of-domain supplementation (\texttt{fresh});
(iii)~two \textbf{fresh-start} $r{=}16, \alpha{=}32$ runs trained from the base model after stabilizing gradient accumulation to effective batch size~$8$ (reducing gradient norm from $3$--$10$ to $1.1$--$2.4$): \texttt{r16} (2,810 mixed pairs) and \texttt{verified} (5,025 pairs, of which 2,794 were generated from 1,500 real course-corpus chunks and passed an LLM factuality verification gate with $64.3\%$ acceptance and $97.7\%$ 5/5 grounding).

\paragraph{Evaluation Suites.}
We cross-reference three evaluation layers:
(1)~\emph{Open-ended proxies} on holdout QA sets ($N{=}176$ and $N{=}120$ weak probes) using char-F1, ROUGE-L, and 1--5 LLM rubric accuracy;
(2)~\emph{Position-swapped pairwise Arena} (\texttt{deepseek-flash}) across 7 comparisons ($N \in \{40, 144, 176\}$), where each pair $(A,B)$ is judged in both orders and order-inconsistent items are logged as position-sensitive flips ($\rho$);
(3)~\emph{Objective Zero-Noise Anchor (New)}: 1,359 4-way Chinese CS/AI multiple-choice questions comprising 5 C-Eval test sets (Computer Architecture 193, Operating Systems 179, Computer Networks 171, Discrete Math 153, Probability \& Statistics 166; subtotal 862) and 3 CMMLU test sets (Computer Science 204, Computer Security 171, Machine Learning 122; subtotal 497).
A character 4-gram Jaccard audit against all 9,991 unique training instructions confirms strict non-contamination (max Jaccard $0.3636$; zero items $\ge 0.50$, only $1$ item $\ge 0.35$).
Each model is scored in a single forward pass (\texttt{logits\_to\_keep=1}) over option tokens \texttt{\{A,B,C,D\}}, yielding exact accuracy, 95\% Wilson CIs~\cite{wilson1927probable}, paired exact McNemar $p$-values~\cite{mcnemar1947note}, 4-way negative log-likelihood (NLL), 10-bin Expected Calibration Error (ECE)~\cite{guo2017calibration}, and raw vocabulary probability mass on \texttt{\{A,B,C,D\}} (\texttt{OptMass}) to measure format-following drift.

\begin{table*}[t]
\centering
\small
\setlength{\tabcolsep}{4.2pt}
\resizebox{\textwidth}{!}{%
\begin{tabular}{llcccccccc}
\toprule
\textbf{Model} & \textbf{Regime (Rank / Eff. BS)} & \textbf{Acc (\%)} & \textbf{C-Eval} & \textbf{CMMLU} & \textbf{NLL $\downarrow$} & \textbf{ECE $\downarrow$} & \textbf{OptMass} & \textbf{$p$ vs Base} & \textbf{$p$ vs Prod} \\
\midrule
\texttt{base} & Unadapted \texttt{Qwen2.5-7B} & \textbf{70.86} & \textbf{64.04} & 82.70 & 2.0309 & 0.2424 & \textbf{100.0\%} & --- & \textbf{0.0100} \\
\midrule
\texttt{init} & Chained ($r{=}8$, bs$=1$, 1ep) & 68.65 & 61.02 & 81.89 & \textbf{0.8600} & \textbf{0.1043} & 90.0\% & 0.0535 & 0.3909 \\
\texttt{continue} & Chained ($r{=}8$, bs$=1$, 3ep) & 68.43 & 61.48 & 80.48 & 1.0022 & 0.1538 & 86.5\% & \textbf{0.0391}$\downarrow$ & 0.4188 \\
\texttt{weak\_focus} & Chained ($r{=}8$, bs$=1$, $71{\times}3$) & \textcolor{red!70!black}{\textbf{67.11}} & 59.98 & 79.48 & 1.0240 & 0.1690 & \textcolor{red!70!black}{85.1\%} & \textbf{0.0012}$\downarrow$ & 0.1524 \\
\texttt{weak\_1to1} & Chained ($r{=}8$, bs$=1$, 1:1 mix) & 67.99 & 60.90 & 80.28 & 1.0205 & 0.1605 & 87.7\% & \textbf{0.0126}$\downarrow$ & 1.0000 \\
\texttt{corr} (\textbf{Prod}) & Chained ($r{=}8$, bs$=1$, corr mix) & \textcolor{red!70!black}{\textbf{67.92}} & 60.67 & 80.48 & 0.9908 & 0.1552 & 90.8\% & \textbf{0.0100}$\downarrow$ & --- \\
\texttt{concept\_v2} & Chained ($r{=}8$, bs$=1$, contrast) & 67.48 & 60.44 & 79.68 & 1.0209 & 0.1643 & 88.8\% & \textbf{0.0032}$\downarrow$ & 0.3269 \\
\texttt{fresh} & Chained ($r{=}8$, bs$=1$, clean OOD) & 67.77 & 60.44 & 80.48 & 1.0050 & 0.1568 & 86.9\% & \textbf{0.0083}$\downarrow$ & 0.8450 \\
\midrule
\texttt{r16} & Fresh ($r{=}16$, bs$=8$, 2.8k) & 69.98 & 62.18 & \textbf{83.50} & 1.4251 & 0.2207 & \textbf{99.9\%} & 0.2354 & 0.0672 \\
\texttt{verified} & Fresh ($r{=}16$, bs$=8$, 5k gated) & \textcolor{teal!80!black}{\textbf{70.79}} & \textbf{63.46} & \textbf{83.50} & 1.2441 & 0.2003 & \textbf{99.6\%} & \textbf{1.0000} & \textcolor{teal!80!black}{\textbf{0.0084}}$\uparrow$ \\
\bottomrule
\end{tabular}}
\caption{\textbf{Objective Zero-Noise CS/AI Benchmark Results ($N{=}1{,}359$; 862 C-Eval + 497 CMMLU).} All six post-init chained $r{=}8$, bs$=1$ models suffer statistically significant catastrophic forgetting ($p \le 0.039$) and format-compliance drift (\texttt{OptMass} drops to $85.1\%$--$90.8\%$). Fresh-start $r{=}16$, bs$=8$ with grounded verification gating (\texttt{verified}) completely eliminates forgetting ($70.79\%$ vs.\ base $70.86\%$, $p{=}1.0$) and significantly outperforms the judge-selected production model \texttt{corr} ($+2.87$\,pp, exact McNemar $p{=}0.0084$).}
\label{tab:checkpoints}
\end{table*}

\section{Proxy Metric and Retrieval Illusions}
\label{sec:proxies}

Before introducing the objective ground-truth audit, three internal evaluation inversions already emerged inside the project's own logs:

\paragraph{1. Validation Loss vs.\ Generation Inversion.}
During 3-epoch continuation (\texttt{continue}), validation loss reached its minimum at epoch~1 ($1.5587$) and worsened monotonically through epoch~2 ($1.5652$) and epoch~3 ($1.6216$), signaling token-level overfitting.
Yet generation char-F1 improved monotonically over the same trajectory ($0.4320 \to 0.4425 \to 0.4539$, vs.\ $0.292$ for \texttt{base}).
Selecting checkpoints purely by validation loss would have stopped at epoch~1, whereas selecting purely by char-F1 invites verbosity overfitting, as seen next.

\paragraph{2. Verbosity/Template Overfitting Under Surface Metrics.}
On 120 held-out weak-point probes, \texttt{weak\_focus} (pure targeted SFT, $71{\times}3$ steps) achieved the highest char-F1 ($0.5435$ vs.\ $0.4579$ baseline) and highest probe judge score ($2.68$ vs.\ $2.35$).
However, on the 60-item general holdout set, its char-F1 ticked up ($0.4459 \to 0.4598$) while its rubric judge accuracy \emph{dropped} ($3.70 \to 3.48$) and average response length swelled by $+19.2$ characters ($123.8 \to 143.0$).
Our objective audit in Section~\ref{sec:objective} confirms that this exact checkpoint suffered the worst factual degradation of all 10 models ($-3.75$\,pp, $p{=}0.0012$).

\begin{figure*}[t]
\centering
\includegraphics[width=0.88\textwidth]{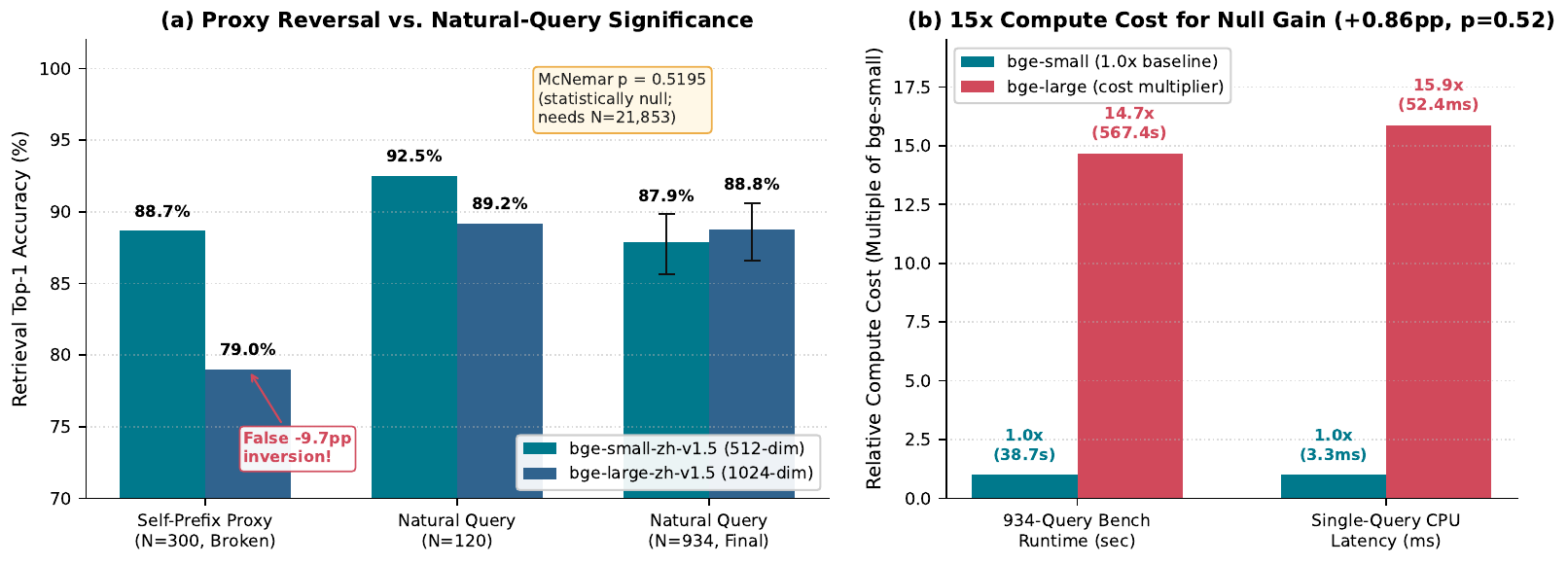}
\caption{\textbf{Retrieval Proxy Inversion and Efficiency Cost.} (a) Self-prefix querying falsely penalizes \texttt{bge-large} by $-9.7$\,pp; scaling to 934 natural student queries yields a statistically null gap (McNemar $p{=}0.5195$). (b) Doubling dimensionality incurs $14.7\times$--$15.9\times$ compute overhead without significant retrieval gain.}
\label{fig:embed}
\end{figure*}

\paragraph{3. Retrieval Proxy Reversal and Zero-Gain Compute Overhead.}
Migrating a 24,170-chunk PostgreSQL/\texttt{pgvector} knowledge base from \texttt{bge-small-zh-v1.5} (512-dim) to \texttt{bge-large-zh-v1.5} (1024-dim)~\cite{xiao2024cpack} exposed a striking benchmark-design trap (Figure~\ref{fig:embed}).
An initial self-retrieval proxy---querying each chunk using its own first 60 characters across 1,915 chunks ($N{=}300$)---ranked \texttt{bge-large} \emph{9.7\,pp worse} than \texttt{bge-small} ($79.0\%$ vs.\ $88.7\%$ Top-1) because lexical self-overlap favors smaller token-matching representations.
Replacing this proxy with 934 LLM-simulated natural student queries against 40 distractors eliminated the inversion, but revealed that the two encoders are statistically indistinguishable: \texttt{bge-large} achieves $88.76\%$ Top-1 (95\% CI $[86.57\%, 90.63\%]$, MRR $0.9313$) versus \texttt{bge-small}'s $87.90\%$ (95\% CI $[85.65\%, 89.84\%]$, MRR $0.9273$), with exact McNemar $p{=}0.5195$ (63 vs.\ 55 discordant queries).
Confirming this $+0.86$\,pp difference at $80\%$ power would require $N{=}21{,}853$ queries, yet \texttt{bge-large} incurs a \textbf{$14.7\times$ benchmark runtime cost} ($567.4$\,s vs.\ $38.7$\,s) and \textbf{$15.9\times$ single-query CPU latency} ($52.4$\,ms vs.\ $3.3$\,ms) for double the vector storage.

\section{Contamination Inflation vs.\ Global Dilution}
\label{sec:contam}

Error-driven synthetic augmentation introduces two opposing distortions simultaneously:

\paragraph{Global Average Dilution.}
After diagnosing 39 low-scoring questions ($22\%$) in the 176-item validation set and training \texttt{corr} on 231 targeted correction samples mixed with general data, evaluating on all 176 items produced an apparent stalemate ($69{:}71$, sign-test $p{=}0.9327$).
Disaggregating the 39 targeted failure items revealed a $22{:}9$ improvement ($p{=}0.029$) that had been washed out by the 137 already-passing items.

\paragraph{Quantifying Data-to-Eval Contamination.}
Auditing \texttt{qa\_shard\_correction.jsonl} (234 entries) against \texttt{qa\_eval\_holdout\_v2.jsonl} (176 entries) using our 4-gram Jaccard auditor (\texttt{sft-audit}) exposed the flip side: generating corrections directly from validation failures leaked $20$ items at Jaccard $\ge 0.35$, $12$ items at $\ge 0.50$, and $3$ verbatim copies ($\text{Jaccard}=1.0$, covering DPO loss code, linearizability vs.\ sequential consistency, and PyTorch variational regression).
Constructing a strictly decontaminated 144-item benchmark across six previously unseen CS subfields (4-gram Jaccard $<0.35$ against all 23,991 corpus items) reduced the decisive win rate of \texttt{prod} over \texttt{init} from $75.0\%$ ($24{:}8$, $p{=}0.007$) on the contaminated weak probe to $69.0\%$ ($87{:}39$, $p<0.0001$) on the clean out-of-domain benchmark---cleanly separating $+6.0$\,pp of contamination inflation from genuine open-ended generalization.

\section{Objective Ground-Truth Audit: Hidden Forgetting and the Rejected Champion}
\label{sec:objective}

\begin{figure*}[t]
\centering
\includegraphics[width=\textwidth]{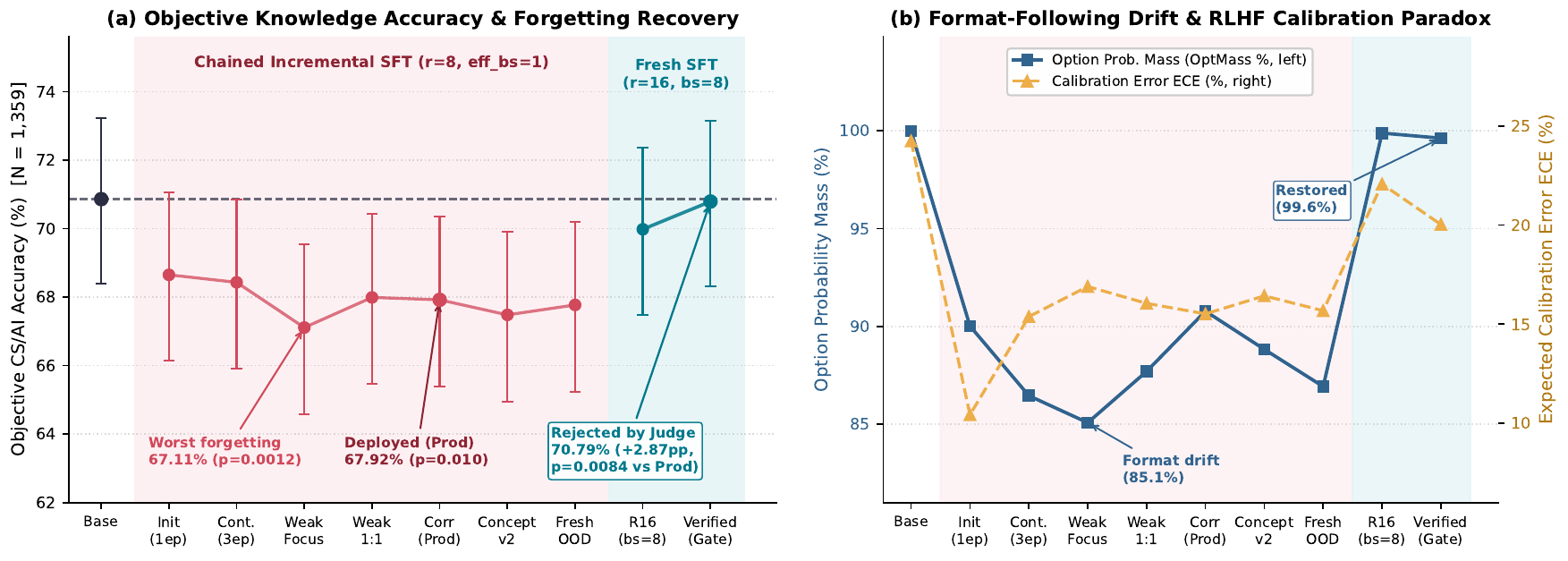}
\caption{\textbf{Ten-Checkpoint Objective Audit.} (a) Chained $r{=}8$, effective-batch-1 fine-tuning causes statistically significant forgetting ($70.86\% \to 67.11\%$--$68.43\%$), whereas fresh-start $r{=}16$, effective-batch-8 retraining with grounded verification gating (\texttt{verified}) completely recovers base accuracy ($70.79\%$) and beats the deployed model \texttt{corr} by $+2.87$\,pp ($p{=}0.0084$). (b) Chained runs erode MCQ format compliance (\texttt{OptMass} drops to $85.1\%$), which fresh-start training restores ($99.6\%$--$99.9\%$), while mild SFT initially softens RLHF overconfidence (ECE $24.2\% \to 10.4\%$) before multi-round overfitting rebounds.}
\label{fig:traj}
\end{figure*}

Running all 10 checkpoints on the 1,359-question uncontaminated C-Eval + CMMLU suite resolves the central ambiguity of the project's five failed optimization rounds---and reverses its deployment decision (Table~\ref{tab:checkpoints} and Figure~\ref{fig:traj}).

\subsection{Chained Incremental SFT Causes Statistically Significant Forgetting}
While open-ended char-F1 rose from $0.292$ (\texttt{base}) to $0.506$ (\texttt{corr}) and pairwise judging awarded \texttt{prod} a $69\%$ win rate over \texttt{init}, objective domain accuracy tells the opposite story:
\begin{itemize}
  \item Every chained run beyond 1 epoch loses between $-2.43$\,pp and $-3.75$\,pp relative to \texttt{base} ($70.86\%$), all statistically significant under exact two-sided McNemar tests ($p \in [0.0012, 0.0391]$).
  \item The deployed production model \texttt{corr} drops to $67.92\%$ ($-2.94$\,pp, $p{=}0.0100$), and the char-F1 leader \texttt{weak\_focus} drops to $67.11\%$ ($-3.75$\,pp, $p{=}0.0012$).
  \item \textbf{Format-Following Drift.} Total softmax probability mass on valid answer letters \texttt{\{A,B,C,D\}} drops from $100.0\%$ (\texttt{base}) to $85.1\%$--$90.8\%$ across chained runs, indicating that repeated batch-1 fine-tuning on 120-character textbook prose steadily overwrites concise instruction-following priors.
\end{itemize}

\subsection{Subject-Level Mechanism and Representational Drift}
\begin{figure*}[t]
\centering
\includegraphics[width=0.92\textwidth]{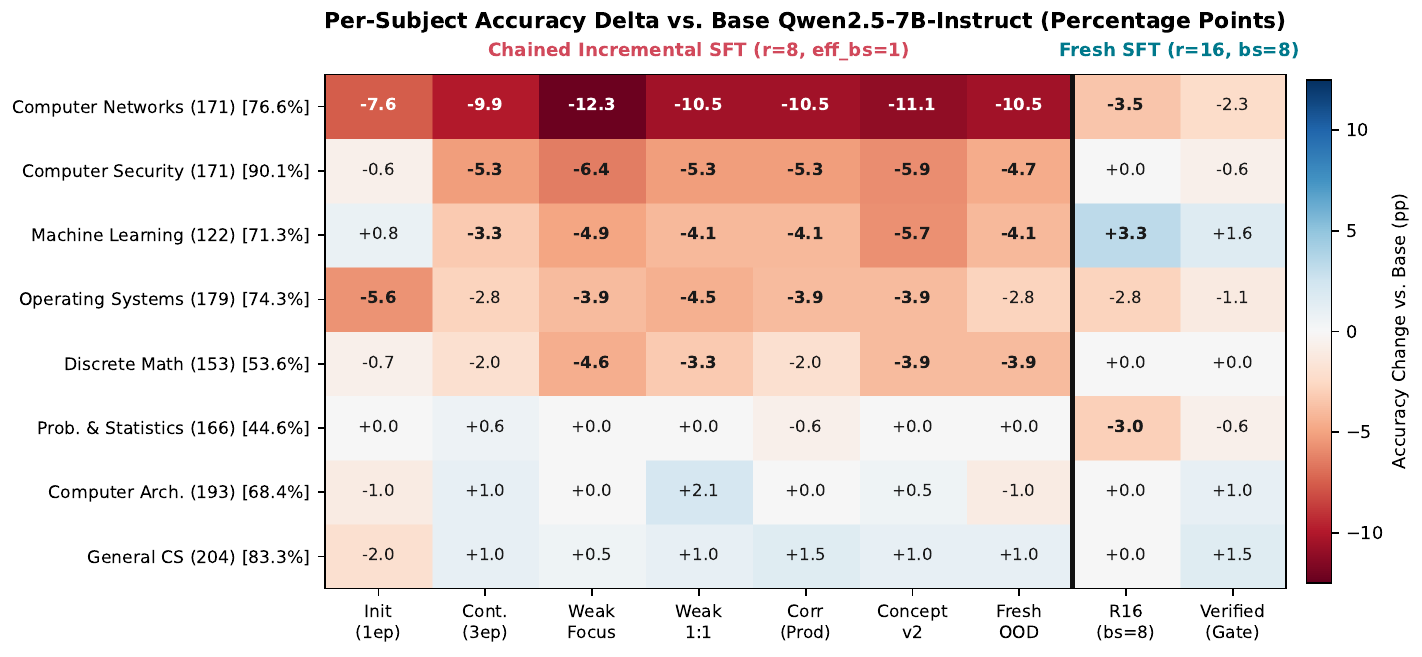}
\caption{\textbf{Per-Subject Accuracy Delta vs.\ Base (\texttt{Qwen2.5-7B-Instruct}).} Chained $r{=}8$, bs$=1$ adapters suffer severe forgetting in Computer Networks (up to $-12.3$\,pp), Security ($-6.4$\,pp), and Machine Learning ($-5.7$\,pp). Fresh-start $r{=}16$, bs$=8$ retraining (\texttt{r16}, \texttt{verified}) sharply reverses these losses and surpasses the base model on Machine Learning ($+1.6$ to $+3.3$\,pp) and General CS ($+1.5$\,pp).}
\label{fig:heatmap}
\end{figure*}

Figure~\ref{fig:heatmap} disaggregates accuracy deltas across all eight subjects.
Forgetting is highly non-uniform and hits both auxiliary and \emph{core in-domain} subjects:
\begin{itemize}
  \item \textbf{Computer Networks (171 items):} collapses by $-10.5$\,pp in \texttt{corr} ($76.6\% \to 66.1\%$) and $-12.3$\,pp in \texttt{weak\_focus} ($64.3\%$), directly explaining why the open-ended clean audit flagged Networks as a weak subfield.
  \item \textbf{Computer Security (171 items):} drops $-5.3$\,pp to $-6.4$\,pp ($90.1\% \to 83.6\%$--$84.8\%$).
  \item \textbf{Machine Learning (122 items):} despite being the largest single subject in the training set (644+ QA pairs), chained SFT \emph{degrades} ML accuracy by $-4.1$\,pp in \texttt{corr} ($71.3\% \to 67.2\%$) and $-5.7$\,pp in \texttt{concept\_v2} ($65.6\%$).
\end{itemize}

\subsection{Fresh-Start $r{=}16$, Batch-8 + Verification Gating Eliminates Forgetting---Yet Was Rejected}
Switching from chained $r{=}8$, bs$=1$ updates to fresh-start $r{=}16$, effective-batch-8 training triggers a sharp V-shaped recovery (Table~\ref{tab:checkpoints}, bottom two rows):
\begin{itemize}
  \item \textbf{Zero Catastrophic Forgetting:} \texttt{verified} achieves $70.79\%$ overall accuracy ($p{=}1.000$ vs.\ \texttt{base} $70.86\%$), restores \texttt{OptMass} to $99.6\%$, and surpasses \texttt{base} on CMMLU ($83.50\%$ vs.\ $82.70\%$) and Machine Learning ($73.0\%$ vs.\ $71.3\%$).
  \item \textbf{Significant Superiority Over Deployed Production:} On the 1,359 paired questions, \texttt{verified} beats the deployed \texttt{corr} model on $124$ items vs.\ $85$ losses ($+2.87$\,pp, exact McNemar $p{=}0.0084$).
  \item \textbf{57\% Reduction in Representational Drift:} Counting total discordant predictions against \texttt{base}, \texttt{corr} flips $230/1{,}359$ answers ($16.9\%$ drift: 135 base-only vs.\ 95 corr-only), whereas \texttt{verified} flips only $99/1{,}359$ ($7.3\%$ drift: 50 vs.\ 49)---less than half the boundary perturbation while achieving the same open-ended textbook style ($61{:}61$ vs.\ \texttt{corr}).
  \item \textbf{The RLHF Calibration Paradox:} \texttt{base} exhibits severe RLHF overconfidence ($\text{ECE}{=}0.2424$, $\text{NLL}{=}2.0309$). One epoch of SFT (\texttt{init}) halves calibration error ($\text{ECE}{=}0.1043$, $\text{NLL}{=}0.8600$) by softening peaked logits, but multi-round chained overfitting pushes ECE back up ($0.1690$). Crucially, \texttt{verified} retains base accuracy while still slashing NLL by $39\%$ ($2.0309 \to 1.2441$).
\end{itemize}

Because \texttt{verified} matches \texttt{corr} on open-ended textbook QA ($61{:}61$), strictly dominates \texttt{corr} on objective CS accuracy ($+2.87$\,pp, $p{=}0.0084$), and restores instruction-following compliance ($99.6\%$ vs.\ $90.8\%$), \texttt{verified} is \textbf{strictly Pareto-superior}.
Yet during the original engineering workflow, the pairwise LLM judge reported an exact $61{:}61$ ($p{=}1.0$) tie, leading the team to discard \texttt{verified} and keep the degraded \texttt{corr} adapter in production.

\section{The Judge Resolution Ceiling: Statistical Theory and Lookup Table}
\label{sec:mde}

\begin{figure*}[t]
\centering
\includegraphics[width=\textwidth]{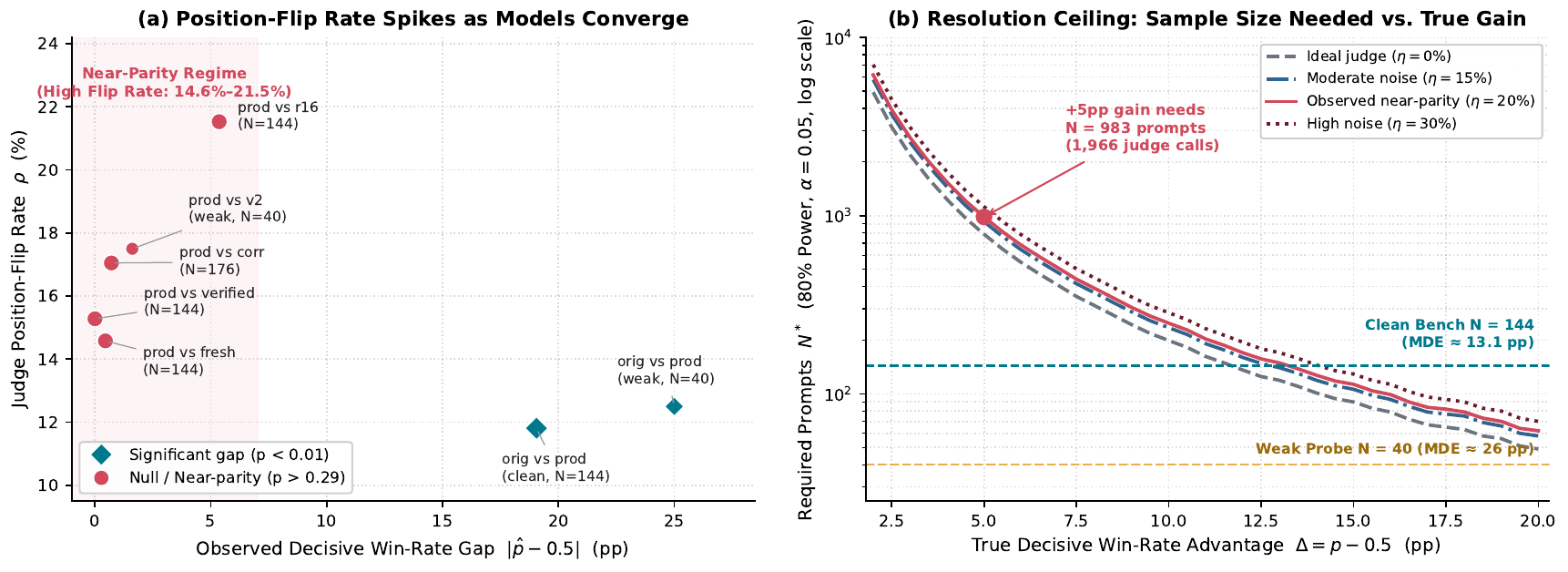}
\caption{\textbf{The Judge Resolution Ceiling.} (a) Empirical Arena runs show position-flip rates climbing from $11.8\%$--$12.5\%$ (large gap) to $14.6\%$--$21.5\%$ near parity. (b) Exact binomial power curves reveal that typical $N{=}40$--$144$ LLM-judge benchmarks sit deep inside a statistical blind zone for incremental gains ($+2$ to $+8$\,pp).}
\label{fig:mde}
\end{figure*}

Why did a carefully designed, position-swapped LLM judge fail to distinguish \texttt{verified} and \texttt{r16} from \texttt{corr}?
Auditing all 7 Arena logs (Table~\ref{tab:arena} and Figure~\ref{fig:mde}) reveals a structural statistical barrier.

\begin{table}[t]
\centering
\scriptsize
\setlength{\tabcolsep}{3.0pt}
\resizebox{\linewidth}{!}{%
\begin{tabular}{lrrrrrrr}
\toprule
\textbf{Comparison} & $N$ & $N_{\text{eff}}$ & \textbf{A:B} & $\rho$ & $\eta$ & \textbf{MDE} & \textbf{Pow(5)} \\
\midrule
\texttt{orig} vs \texttt{prod} (weak) & 40 & 32 & 8:24 & 12.5\% & 20.0\% & 26.5\% & 4.1\% \\
\texttt{orig} vs \texttt{prod} (clean) & 144 & 126 & 39:87 & 11.8\% & 12.5\% & 12.7\% & 17.7\% \\
\midrule
\texttt{prod} vs \texttt{corr} (all) & 176 & 140 & 69:71 & 17.0\% & 20.5\% & 12.4\% & 17.6\% \\
\texttt{prod} vs \texttt{v2} (weak) & 40 & 31 & 15:16 & 17.5\% & 22.5\% & 25.7\% & 5.5\% \\
\texttt{prod} vs \texttt{fresh} (clean) & 144 & 111 & 56:55 & 14.6\% & 22.9\% & 13.7\% & 15.0\% \\
\texttt{prod} vs \texttt{ver} (clean) & 144 & 122 & 61:61 & 15.3\% & 15.3\% & 13.1\% & 16.4\% \\
\texttt{prod} vs \texttt{r16} (clean) & 144 & 112 & 62:50 & \textbf{21.5\%} & 22.2\% & 13.2\% & 17.7\% \\
\bottomrule
\end{tabular}}
\caption{\textbf{Empirical Resolution Audit of All 7 Position-Swapped Arena Runs.} $\rho$: position-flip rate; $\eta$: total indecisive rate ($\rho+\text{tie}$); MDE: exact 80\%-power Minimum Detectable decisive advantage ($\alpha{=}0.05$); Pow(5): statistical power to detect a true $+5$\,pp gain.}
\label{tab:arena}
\end{table}

\paragraph{Position-Flip Inflation Near Parity.}
When candidate quality differs strongly (\texttt{orig} vs.\ \texttt{prod}, win-share $69\%$--$75\%$), judge position-flip rates stay moderate ($\rho = 11.8\%$--$12.5\%$).
Once iterative SFT brings candidates close in open-ended style (win-share $50\%$--$55\%$), $\rho$ climbs to $14.6\%$--$21.5\%$ (Pearson $r = -0.599$ between gap $|\hat{p}-0.5|$ and $\rho$), and combined indecisive rates $\eta = \rho + \tau$ reach $15.3\%$--$22.9\%$.
This behavior follows directly from any latent-margin judge model with position bias $\gamma$~\cite{wang2024large}: when the true quality margin $d_i$ approaches $0$, the probability that $d_i - \gamma + \epsilon_1$ and $d_i + \gamma + \epsilon_2$ disagree attains its maximum.

\paragraph{Exact Minimum Detectable Effect (MDE).}
Under position-swapped judging on $N$ prompts, dropping ties and order flips leaves $N_{\text{eff}} = (1 - \eta) N$ decisive pairs.
For a two-sided exact binomial sign test of $H_0: p{=}0.5$ against $H_1: p{=}0.5+\Delta$ at level $\alpha$ and power $1-\beta$, the required total prompt budget scales as:
\begin{equation}
  N^*(\Delta, \eta) \approx \frac{1}{1 - \rho - \tau} \left( \frac{0.5\, z_{\alpha/2} + z_\beta \sqrt{0.25 - \Delta^2}}{\Delta} \right)^2.
  \label{eq:nreq}
\end{equation}
As candidates converge, iterative SFT suffers a \emph{double penalty}: $\Delta$ shrinks quadratically in the denominator while $\rho$ increases, inflating $N^*$ further.
Evaluating exact binomial inversion (Table~\ref{tab:lookup} and Figure~\ref{fig:mde}b) shows that:
\begin{itemize}
  \item At $N{=}144$ and $\eta \in [15\%, 23\%]$, the 80\%-power MDE is \textbf{$13.1$--$13.7$\,pp} (requiring a $63\%+$ vs.\ $37\%$ decisive landslide). Power for a realistic $+5$\,pp gain is only \textbf{$15.0\%$--$17.7\%$}.
  \item Reliably detecting a $+5$\,pp decisive improvement at $\eta{=}20\%$ requires \textbf{$N{=}983$ prompts} ($1{,}966$ swapped judge calls); detecting $+3$\,pp requires \textbf{$N{=}2{,}737$ prompts} ($5{,}474$ judge calls).
\end{itemize}

\begin{table}[t]
\centering
\small
\setlength{\tabcolsep}{3.8pt}
\resizebox{\linewidth}{!}{%
\begin{tabular}{lrrrrr}
\toprule
\textbf{True Gain $\Delta$} & $\eta{=}0\%$ & $\eta{=}10\%$ & $\eta{=}15\%$ & $\eta{=}20\%$ & $\eta{=}30\%$ \\
\midrule
$+15$\,pp ($50{\to}65\%$) & 90 & 100 & 106 & 113 & 129 \\
$+10$\,pp ($50{\to}60\%$) & 199 & 222 & 235 & 249 & 285 \\
$+8$\,pp ($50{\to}58\%$) & 312 & 347 & 368 & 390 & 446 \\
$+5$\,pp ($50{\to}55\%$) & 786 & 874 & 925 & \textbf{983} & 1,123 \\
$+3$\,pp ($50{\to}53\%$) & 2,189 & 2,433 & 2,576 & \textbf{2,737} & 3,128 \\
$+2$\,pp ($50{\to}52\%$) & 4,927 & 5,475 & 5,797 & 6,159 & 7,039 \\
\bottomrule
\end{tabular}}
\caption{\textbf{Practitioner Sample-Size Lookup Table.} Minimum prompts $N^*$ (each requiring $2$ swapped judge calls) needed for $80\%$ power ($\alpha{=}0.05$) under exact binomial sign testing at indecisive rate $\eta = \rho + \tau$.}
\label{tab:lookup}
\end{table}

\section{The \texttt{sft-audit} Four-Gate Protocol}
\label{sec:protocol}

To make these diagnostics immediately reusable by low-resource teams, we package our methodology into a standalone CPU-only CLI (\texttt{sft-audit}) enforcing four sequential gates before any model or retriever promotion:
\begin{enumerate}
  \item \textbf{Gate 1 --- Contamination Guard (\texttt{decontaminate}):} Character 4-gram Jaccard inverted-index scan across all training shards vs.\ held-out evaluation sets; blocks promotion or flags inflation whenever any item exceeds $0.35$.
  \item \textbf{Gate 2 --- Judge Power Check (\texttt{mde}):} Computes the exact MDE of a pairwise judge run given $N$ and observed $\eta$; explicitly distinguishes \emph{evidence of absence} from \emph{underpowered noise}.
  \item \textbf{Gate 3 --- Objective Ground-Truth Anchor (\texttt{mcnemar}):} Evaluates candidates on $\ge 1{,}000$ zero-noise domain MCQs via single-forward option logits ($\sim 3$\,min per 7B adapter on an 8\,GB GPU), reporting exact McNemar $p$-values and discordant-pair drift against both \texttt{base} and the incumbent production checkpoint.
  \item \textbf{Gate 4 --- Format \& Cost Hygiene:} Tracks \texttt{OptMass} (flagging format drift $<95\%$), NLL/ECE calibration, and compute-adjusted significance for retriever upgrades (avoiding $15\times$ latency penalties for McNemar-null gains).
\end{enumerate}

\paragraph{Practical Takeaway for 8\,GB QLoRA Practitioners.}
Our 10-checkpoint autopsy yields a concrete training rule: \textbf{avoid chaining multiple sequential \texttt{batch=1} synthetic SFT patches on top of an already-adapted LoRA}.
When new verified data is collected, \emph{retrain fresh from the base model} with gradient accumulation ($\text{eff\_bs} \ge 8$) and higher rank ($r \ge 16$), and gate deployment jointly on an objective anchor plus open-ended checks.

\section{Related Work}
\paragraph{LLM-as-a-Judge Biases.}
Prior work established position bias, verbosity preference, and self-enhancement in LLM evaluators~\cite{zheng2023judging,wang2024large}.
Our contribution connects position-flip inflation near model parity directly to exact binomial statistical power, showing empirically how a $15.3\%$--$21.5\%$ flip floor causes real-world rejection of a statistically superior model.

\paragraph{Catastrophic Forgetting and Calibration in SFT.}
Continual fine-tuning of LLMs is known to induce alignment tax and general-capability forgetting~\cite{luo2025forgetting}, while RLHF models exhibit post-alignment miscalibration~\cite{guo2017calibration}.
We complement large-scale studies by isolating the exact interaction between chained \texttt{batch=1} QLoRA, synthetic error-correction leakage, format-compliance decay (\texttt{OptMass}), and fresh-start recovery under strict 8\,GB consumer hardware constraints.

\section{Limitations}
Our study examines a single base model family (\texttt{Qwen2.5-7B-Instruct}) and one domain (Chinese computer science education), with pairwise judging performed by a single judge model (\texttt{deepseek-flash}).
While multi-judge ensembles can reduce $\rho$, Equation~\eqref{eq:nreq} applies identically to any position-swapped judge once its empirical indecisive rate $\eta$ is measured.
Finally, our objective anchor measures closed-form factual knowledge and format compliance via MCQ logits; it complements---rather than replaces---human evaluation of open-ended pedagogical clarity.

\section{Conclusion}
Iterative low-resource fine-tuning fails silently at both ends of the evaluation pipeline: naive proxies reward style overfitting while hiding significant knowledge forgetting, and underpowered pairwise LLM judges discard genuine structural fixes once models converge into the judge's position-flip noise floor.
By combining a zero-noise objective anchor ($N{=}1{,}359$) with exact power auditing, we uncovered a $+2.87$\,pp ($p{=}0.0084$) Pareto-superior checkpoint that standard LLM judging had misclassified as a $61{:}61$ tie, and demonstrated that fresh-start, stabilized-batch, verification-gated QLoRA completely eliminates catastrophic forgetting on an 8\,GB GPU.

{\small
\setlength{\bibsep}{2.5pt}
\bibliographystyle{plainnat}
\bibliography{references}
}

\end{document}